\documentclass[letterpaper]{article} % DO NOT CHANGE THIS
\usepackage{aaai25}  % DO NOT CHANGE THIS
\usepackage{times}  % DO NOT CHANGE THIS
\usepackage{helvet}  % DO NOT CHANGE THIS
\usepackage{courier}  % DO NOT CHANGE THIS
\usepackage[hyphens]{url}  % DO NOT CHANGE THIS
\usepackage{graphicx} % DO NOT CHANGE THIS
\usepackage{natbib}  % DO NOT CHANGE THIS AND DO NOT ADD ANY OPTIONS TO IT
\usepackage{caption} % DO NOT CHANGE THIS AND DO NOT ADD ANY OPTIONS TO IT
\usepackage{algorithm}
\usepackage{algorithmic}
\usepackage{upgreek}
\usepackage{newfloat}
\usepackage{listings}
\DeclareCaptionStyle{ruled}{labelfont=normalfont,labelsep=colon,strut=off} % DO NOT CHANGE THIS
\floatstyle{ruled}
\newfloat{listing}{tb}{lst}{}
\floatname{listing}{Listing}
\usepackage{booktabs}
\usepackage{array}
\usepackage{multirow}
\usepackage{subfigure}
\usepackage{color}
\usepackage{amsmath}

\title{BitPipe: Bidirectional Interleaved Pipeline Parallelism \\ for Accelerating Large Models Training}
\author{
	Houming Wu, Ling Chen\thanks{Corresponding Author}, Wenjie Yu
}
\affiliations{
	State Key Laboratory of Blockchain and Data Security, College of Computer Science and Technology\\ Zhejiang University, Hangzhou, China \\
\{houmingwu, lingchen, wenjieyu\}@cs.zju.edu.cn
}

\usepackage{bibentry}
\begin{document}
%\iffalse
\maketitle

\begin{abstract}
 With the increasing scale of models, the need for efficient distributed training has become increasingly urgent. Recently, many synchronous pipeline parallelism approaches have been proposed to improve training throughput. However, these approaches still suffer from two major issues, i.e., pipeline bubbles caused by periodic flushing and extra communication due to the increasing number of pipeline stages. To this end, we propose BitPipe, a \underline{\textbf{bi}}directional in\underline{\textbf{t}}erleaved \underline{\textbf{pipe}}line parallelism for accelerating large models training. Specifically, a hybrid scheme of fusing interleaved pipelines with bidirectional pipelines is proposed to reduce the computational time of each single micro-batch and multiply the number of devices executing simultaneously. A V-shaped schedule with eager gradient synchronization is introduced to reduce and overlap the communication between devices. Experiments conducted on up to 32 GPUs show that BitPipe improves the training throughput of GPT-style and BERT-style models by $1.05\times$-$1.28\times$ compared to the state-of-the-art synchronous approaches.% The code of our implementation is available at \url{https://github.com/wuhouming/BitPipe}.
\end{abstract}

% Uncomment the following to link to your code, datasets, an extended version or similar.
%
% \begin{links}
%     \link{Code}{https://aaai.org/example/code}
%     \link{Datasets}{https://aaai.org/example/datasets}
%     \link{Extended version}{https://aaai.org/example/extended-version}
% \end{links}

\section{Introduction}

Scaling the number of parameters in contemporary deep learning models has yielded remarkable the state-of-the-art (SOTA) results. Training these large models is challenging, as the limited memory and computational capacity of a single device (e.g., GPU) pose obstacles to accommodating them within realistic timeframes. For instance, training a GPT-3 175B model demands over 3,000 GiB for storing model parameters and optimizer states, requiring an impractical 288 years with a single  NVIDIA V100 GPU \cite{kim2023bpipe, narayanan2021efficient}. 

The urgency for parallel and distributed training (e.g., data parallelism and model parallelism) has become increasingly pronounced. While data parallelism \cite{li2014scaling} allows for ideal speedup, it falters when confronted with large models that exceed the capacity of a single device. Model parallelism \cite{dean2012large,lee2014model,wang2019supporting} addresses this limitation by distributing the weight parameters of a model across multiple devices, which mitigates the memory usage per device but suffers from severe resource under-utilization. Pipeline parallelism improves resource utilization, which splits a batch into smaller micro-batches and divides a model into stages within a pipeline, allowing simultaneous execution of different micro-batches across multiple devices. Pipeline parallelism can be categorized into synchronous and asynchronous schemes based on weight update semantic. Synchronous approaches flush periodically at the end of each iteration to guarantee strict optimizer semantics, which causes device idle times (also called pipeline bubbles). Asynchronous approaches do away with flushes completely by delaying weight updates, but at the expense of strict model convergence and thus are not within the scope of our work.

\begin{figure}[t]
	\centering
	\includegraphics[width=0.95\columnwidth]{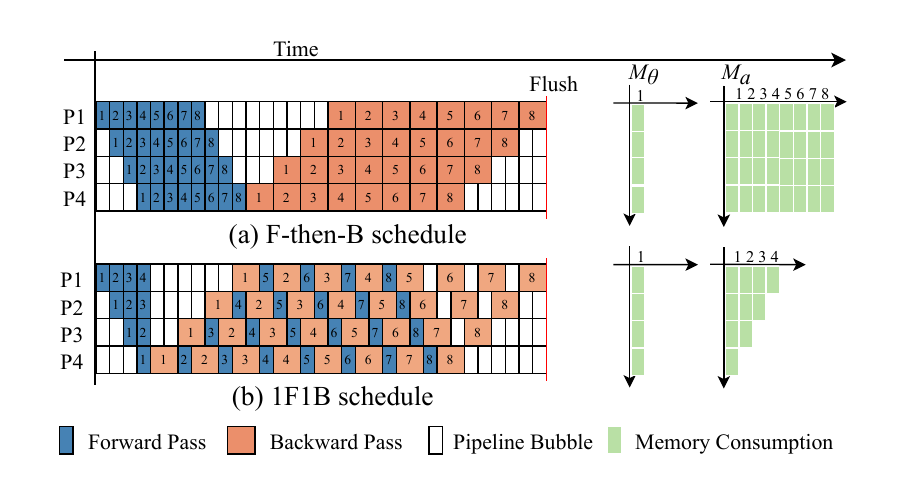} % Reduce the figure size so that it is slightly narrower than the column. Don't use precise values for figure width.This setup will avoid overfull boxes.
	\caption{Classic synchronous pipeline schedules, with 4 pipeline devices and 8 micro-batches within a training iteration. Both schedules have the same bubble overhead and weights memory consumption ($M_\uptheta$). The activations memory consumption ($M_{\rm a}$) of the 1F1B schedule exhibits better efficiency but existing imbalance.}
	\label{defaultschedule}
\end{figure}

Early synchronous approach (e.g., GPipe \cite{huang2019gpipe}) focuses on reducing pipeline bubbles by increasing the number of concurrent batches in the pipeline (as shown in Figure \ref{defaultschedule}(a)). As a direct consequence, there is an increase in peak activation memory demands. Subsequently, encouraged by the success of the 1F1B schedule  (as shown in Figure \ref{defaultschedule}(b)), researchers have proposed memory-efficient approaches (e.g., DAPPLE \cite{fan2021dapple} and PipeDream-Flush \cite{narayanan2021memory}), which further adjusts the number of micro-batches injected into devices at the beginning of pipelines. 

Recently approaches attempt to increase the number of devices executing simultaneously (i.e., bidirectional pipeline parallelism), or to reduce the computational time of a single micro-batch (i.e., interleaved pipeline parallelism), which shows the SOTA performance. In the bidirectional approaches \cite{jain2020gems,li2021chimera,zhang2023mixpipe}, each device stores multiple pipeline stages in different directions, which decreases bubble size and achieves a more balanced activation memory consumption. On the other hand, interleaved approaches \cite{narayanan2021efficient,lamy2023breadth,liu2023hanayo} assign multiple smaller and nonconsecutive stages to each device, which makes each bubble size smaller accordingly.

Despite the promising results, the latest synchronous approaches still face two primary issues. First, the remaining bubbles still pose the largest deficiency. Due to computation dependencies in the pipeline across different devices, bubbles are inevitable. In existing approaches, as much as 50\% of the time can be spent to flush the pipeline. Second, the communication overhead remains considerable even though pipeline parallelism employs point-to-point (P2P) communication. Specifically, bidirectional pipeline parallelism requires additional weight memory and data-parallel communication to reduce pipeline bubbles, while interleaved pipeline parallelism shrinks bubble size at the expense of extra P2P communication. Moreover, if the bidirectional pipeline extends to more than two pipelines, or each device in the interleaved pipeline generalizes to have more stages, the extra communication or memory usage will increase accordingly, further degrading their performance.

To address the aforementioned issues, we propose BitPipe, a bidirectional interleaved pipeline parallelism for accelerating large models training. To the best of our knowledge, BitPipe is the first work that incorporates the interleaved schedule into bidirectional pipeline parallelism, which reduces the computational time of each single micro-batch and doubles the number of devices executing simultaneously. BitPipe transforms the looping schedule of the interleaved pipeline to a V-shaped schedule and thus mitigates the side effect of the additional communication overhead. The contributions of BitPipe are summarized as follows:
	%好处叠加了，坏处如何处理？
\begin{itemize}
	\item  We propose a hybrid pipeline scheme of fusing interleaved pipelines with bidirectional pipelines. This design can not only improve throughput, but also achieves a harmonious balance in memory utilization.
	\item  We introduce a V-shaped schedule of partially transforming cross-device communication to local copying, alongside an eager gradient synchronization scheme, which can reduce and overlap communication between devices. 
	\item  Experiments show that BitPipe can improve the end-to-end performance by up to $1.28\times$ per iteration for GPT-style and BERT-style models compared to the SOTA synchronous pipeline approaches.
\end{itemize}

\section{Related Work}

\subsection{Model Parallelism} 

Model parallelism is a solution to train large models by partitioning the weight parameters of a model among available devices in two ways: tensor (intra-layer) model parallelism \cite{wang2019supporting,shoeybi2019megatron} and inter-layer model parallelism \cite{2012ImageNet,dean2012large}. The former is trapped in requiring all-to-all communication, while the latter suffers from underutilized resources.

\subsection{Pipeline Parallelism}
 
Pipeline parallelism can effectively improve resource utilization. In this scenario, a batch is further partitioned into smaller micro-batches, which allows each device to commence processing the subsequent micro-batch immediately after completing the preceding one. Pipeline parallelism approaches can be categorized into synchronous and asynchronous schemes based on the weight update semantics. For synchronous approaches, the magnitude of pipeline bubbles can be quantified as \emph{bubble ratio}, which is defined as the bubble overhead divided by the overall pipeline runtime. GPipe \cite{huang2019gpipe} reduces the bubble ratio by increasing the number of concurrent batches in the pipeline, which increases the peak activation memory demands as a direct consequence. DAPPLE \cite{fan2021dapple} and PipeDream-Flush \cite{narayanan2021memory} lower the activation memory usage by adjusting the number of micro-batches injected into devices at the beginning of pipelines and performing the 1F1B schedule. Recently efforts have led to bidirectional pipeline parallelism and interleaved pipeline parallelism.

\begin{figure*}[ht]
	\centering
	\includegraphics[width=0.9\textwidth]{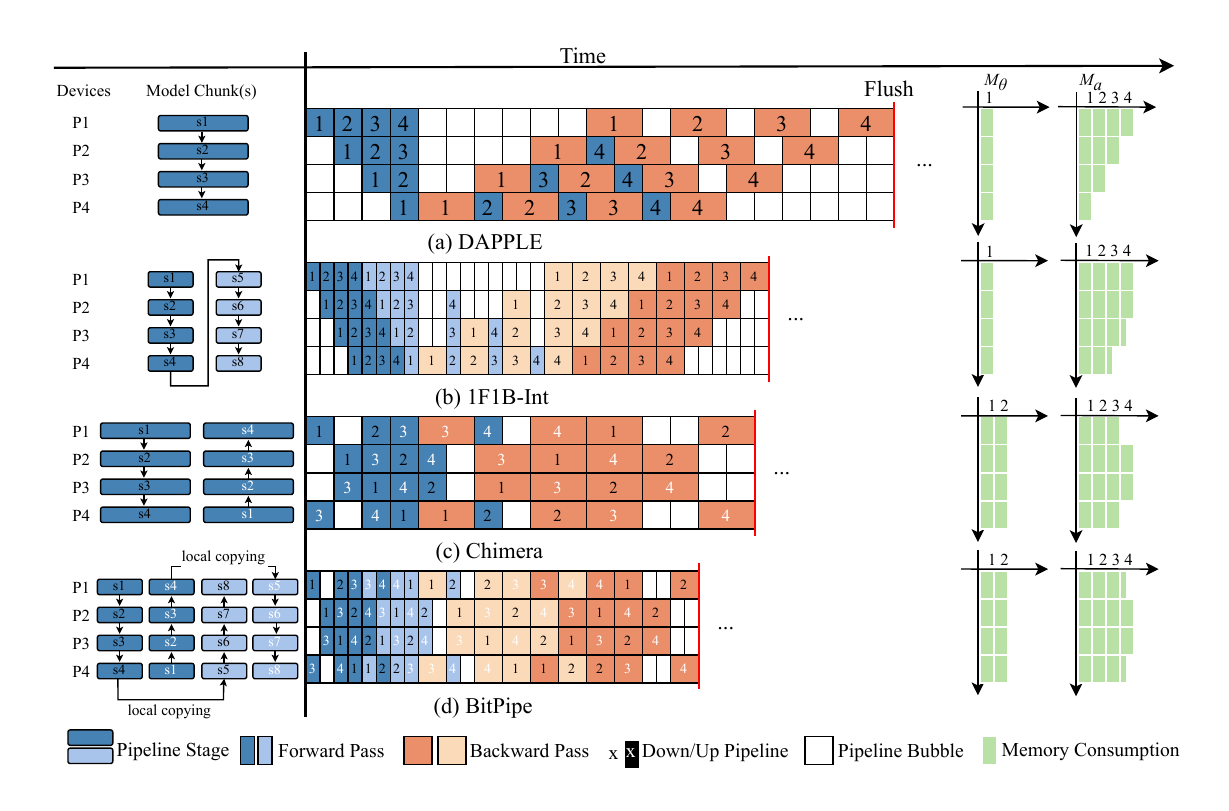}
	\caption{Synchronous approaches considered in this paper, with 4 pipeline devices and 4 micro-batches within a training iteration. Dark colors show the first stage and light colors show the second stage. In Chimera, each device is responsible for 2 pipelines in different directions (black text colors represent the down pipeline and white text colors for the up pipeline).}
	\label{allschedules}
\end{figure*}

\textbf{Bidirectional Pipeline Parallelism} combines two pipelines in different directions, which doubles the number of devices executing simultaneously. GEMS \cite{jain2020gems} is a memory-efficient pipeline approach that first schedules micro-batches among two model replicas. Since GEMS is mainly designed for small batch sizes and executes at most two micro-batches simultaneously, its bubble ratio is much higher than that of the other approaches. Chimera \cite{li2021chimera} implements two pipelines in opposite directions simultaneously (named down and up pipeline, respectively), and the pipeline utilization can be better than that of vanilla pipeline parallelism with a single pipeline, as shown in Figure \ref{allschedules}(c). MixPipe \cite{zhang2023mixpipe} flexibly regulates the number of micro-batches injected into the bidirectional pipelines at the beginning, which achieves a better balance between pipeline utilization and device utilization. However, these approaches impose an increased burden on each device, requiring more weight memory and data-parallel communication.

\textbf{Interleaved Pipeline Parallelism} splits the original pipeline stage into smaller non-consecutive stages and schedules in a loop way, which makes the bubble size reduce with the decrease of the calculation time of each micro-batch. 1F1B-Int \cite{narayanan2021efficient} effectively reduces the bubble ratio without incurring additional memory consumption for model weights, at the cost of extra pipeline-parallel communication overhead. WPipe \cite{ yang2022group} integrates 1F1B-Int with PipeDream-2BW \cite{narayanan2021memory}, which achieves better memory efficiency and fresher weight updates. Breadth-First \cite{lamy2023breadth} generalizes 1F1B-Int and combines it with data parallelism, which shows a better overlap of communication with computation. Hanayo \cite{liu2023hanayo} transforms bidirectional pipeline into a wave-like interleaved pipeline and employs a high-performance execution runtime to enable communication and computation overlap. These studies show the effectiveness of increasing the number of pipeline stages and optimizing communication. 

\begin{figure*}[ht]
	\centering
	\includegraphics[width=0.9\linewidth]{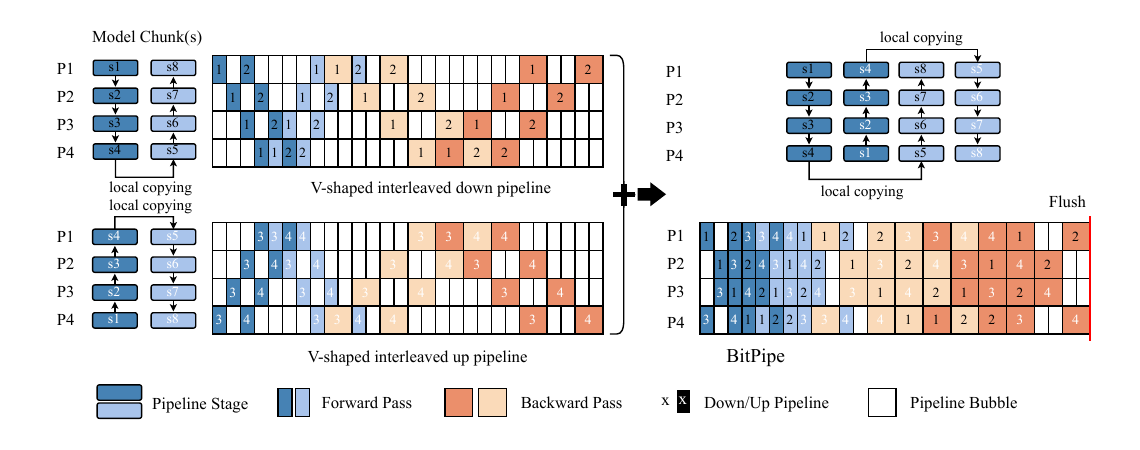}
	\caption{Model chunks and bidirectional interleaved pipelines scheduling of BitPipe, with 4 pipeline devices and 4 micro-batches within a training iteration.}
	\label{BitPipe-V}
\end{figure*}

\section{Methodology}
\subsection{Overview} 
BitPipe is a hybrid schedule of integrating interleaved pipelines with bidirectional pipelines, which makes the bubble ratio smaller and exhibits a more balanced activations memory consumption, as shown in Figure \ref{allschedules}(d). The key idea of BitPipe is to seamlessly merge two V-shaped interleaved pipelines in opposite directions (as shown in Figure \ref{BitPipe-V}), which partially transforms the cross-device communication to local copying and reduces the communication overhead. The symbols used by the following sections are defined in Table \ref{table-symbols}.

\begin{table}[ht]
	\small
	\begin{center}
		\begin{tabular}{ll}
			\toprule
			%Symbols & Definitions \\
			%\midrule
			$D$ & The number of pipeline devices \\
			$W$ & The number of replicated pipelines \\
			$P$ & The number of devices (=$W\ast$$D$)   \\	
			$B$ & Micro-batch size \\
			$N$ & The number of micro-batches in a mini-batch \\
			$\hat{B}$ & Mini-batch size (=$B\ast$$N\ast$$W$)   \\	
			$M_\uptheta$ & Memeory consumption for the weights of one stage \\
			$M_{\rm a}$ & Memeory consumption for the activations of one stage \\
			%$T_{\rm P}$ & The number of samples processed per second \\
			\bottomrule
		\end{tabular}
	\end{center}
	\caption{Symbols description.}
	\label{table-symbols}
\end{table}

\subsection{V-shaped Interleaved Schedule}

\begin{figure}[ht]
	\centering
	\includegraphics[width=0.95\columnwidth]{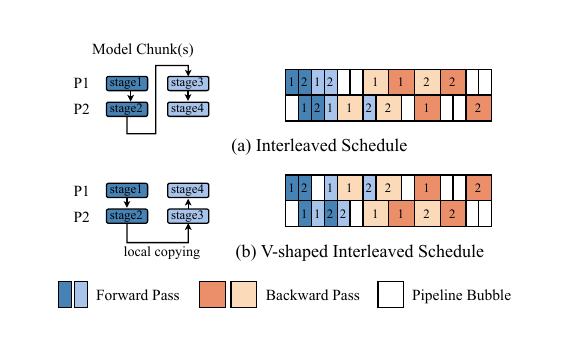}
	\caption{Interleaved pipeline schedules, with 2 pipeline devices and 2 micro-batches for simplicity.}
	\label{Interleaved-V}
\end{figure}

Pipeline parallelism typically splits the model layers into a single stage per device. 1F1B-Int assigns multiple smaller and nonconsecutive stages to each device, with cross-device communication between stages, as illustrated in Figure \ref{allschedules}(b). In contrast to this looping schedule, we introduce the V-shaped interleaved schedule that swaps the order and sequentially allocates stages to devices, starting from the first device and progressing to the last, then reversing the order from the last device back to the first, creating a ``V" shape (i.e., stage1\textasciitilde stage2 are mapped to P1\textasciitilde P2, and stage3\textasciitilde stage4 are mapped to P2\textasciitilde P1, as illustrated in Figure \ref{Interleaved-V}(b)). Since the sequence of computation remains unchanged and the communication overhead is decreased by local copying (between consecutive stages in P2), we can deduce that the efficiency of this V-shaped schedule is at least on a par with, if not superior to, the original looping schedule.

Motivated by Chimera that involves two pipelines and combines them together, we initially contemplate scaling the number of the V-shaped interleaved pipelines in BitPipe. It should be noted that the V-shaped interleaved schedule can be generalized to a greater number of stages while maintaining the unchanged number of pipelines (discussed in Appendix A), and this further reduces the bubbles but at the expense of higher communication overhead.

\begin{table}[t]
	\small
	\centering
	\begin{tabular}{lccc}
		\toprule
		\begin{tabular}[l]{@{}l@{}}
			Pipeline\\Approaches
		\end{tabular} &
		\begin{tabular}[c]{@{}c@{}}
			Bubble\\Ratio 
		\end{tabular} &
		\begin{tabular}[c]{@{}c@{}}
			Weights\\Memory
		\end{tabular} &
		\begin{tabular}[c]{@{}c@{}}
			Activations\\Memory
		\end{tabular} \\
		\midrule
		GPipe &  $\frac{D-1}{N+D-1}$    & $M_\uptheta $&$N\times$$M_{\rm a}$ \\
		DAPPLE &  $\frac{D-1}{N+D-1}$    & $M_\uptheta$ &$[M_{\rm a}$, $D\times$$M_{\rm a}]$  \\
		1F1B-Int & $\frac{D-1}{2N + D-1}$  & $M_\uptheta $&$[\frac{D+1}{2}M_{\rm a}$, $D\times$$M_{\rm a}]$     \\
		Chimera & $\frac{D-2}{3N/2 + D-2}$ & $2M_\uptheta $&$[\frac{D+2}{2}M_{\rm a}$, $D\times$$M_{\rm a}]$           \\
		\textbf{BitPipe} & $\frac{D-2}{3N+D-2}$ & $2M_\uptheta$ &$[\frac{D+3}{2}M_{\rm a}$, $D\times$$M_{\rm a}]$ \\
		\bottomrule
	\end{tabular}
	\caption{Comparison of different pipeline approaches.}
	\label{table-methods}
\end{table}

\subsection{Bidirectional Interleaved Pipelines}\label{sec:bitpipe}

The core concept of BitPipe is seamlessly integrating two V-shaped interleaved pipelines in opposite directions. Figure \ref{BitPipe-V} presents an example with four pipeline devices (i.e., $D$=4). Herein, we assume that each device executes $D$ micro-batches within a training iteration (i.e., $N$=$D$), which is the minimum to maintain the activeness of all stages. In the V-shaped interleaved down pipeline, stage1\textasciitilde stage4 are mapped to P1\textasciitilde P4, and stage5\textasciitilde stage8 are mapped to P4\textasciitilde P1. The stages in the V-shaped interleaved up pipeline are mapped in strikingly opposite order. Each pipeline schedules $N$/2  (assuming $N$ is an even number) micro-batches using 1F1B-Int strategy, as shown in the left part of Figure \ref{BitPipe-V}. Subsequently, by fusing these two pipelines together, we acquire the BitPipe (the right part of Figure \ref{BitPipe-V}). Given an even number of devices $D$, it is guaranteed that there is no conflict (i.e., there is at most one micro-batch occupying the same time slot on each device) during the merging process.

\subsubsection{Bubble Ratio.} For the synchronous approaches, the bubble ratio is defined as the ratio of the bubble overhead to the overall runtime of the pipeline. By counting the number of injected micro-batches on each device of BitPipe in Figure \ref{BitPipe-V}, it can be observed that BitPipe incurs (3$D$-6)/4 bubbles (i.e., {($D$-2)/2} bubbles in the forward passes and ($D$-2)/4 bubbles in the backward passes). Given the assumption that the workload of a backward pass is about two times of a forward pass, the bubble ratio of BitPipe is ($D$-2)/(3$N$+$D$-2). Table \ref{table-methods} presents the bubble ratio of the five pipeline approaches, among which BitPipe is the lowest. The bubble ratio of BitPipe can be further reduced to ($D$-2)/(4$N$+$D$-2) by removing the middle bubbles (detailed in Appendix B).

\begin{figure}[ht]
	\centering
	\includegraphics[width=0.9\columnwidth]{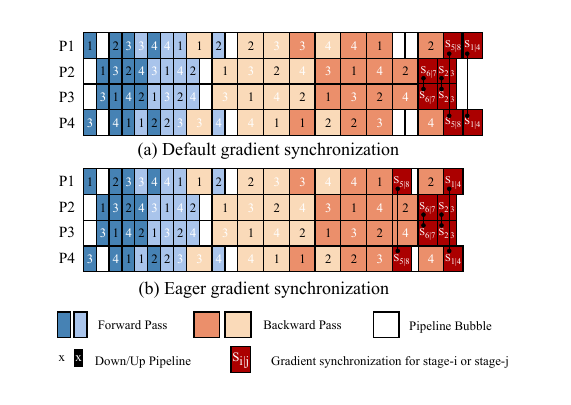}
	\caption{Overlap communication by eager gradient synchronization.}
	\label{EagerSync}
\end{figure}

\subsubsection{Memory Consumption.} Memory consumption is primarily influenced by two aspects: the weight parameters and the intermediate activations. Table \ref{table-methods} also presents the memory usage of BitPipe, where $M_\uptheta$ represents the memory consumption of the weights in one device for one model replica, and $M_{\rm a}$ is the memory consumption of the activations in one device for one micro-batch (as the light green colors shown in Figure \ref{allschedules}). Regarding the weights memory, GPipe, DAPPLE, and 1F1B-Int maintain the weights of one pipeline stage in each device, while Chimera and BitPipe hold two. Concerning the activations memory, GPipe injects $N$ micro-batches into the pipeline concurrently ($N$$\geq$$D$ to fully exploit the pipeline), leading to memory consumption that is proportional to $N$, which does not scale favorably to large mini-batches. Conversely, BitPipe and the other three schedules inject up to $D$ micro-batches at the beginning of the pipeline, which makes memory consumption proportional to  $D$ and scales better.

\subsection{Communication Optimization}

BitPipe applies P2P communication to transfer the intermediate activations and gradients between pipeline stages (except consecutive stages in the same device using local copying). As BitPipe combines bidirectional pipelines together, collective communication (i.e., allreduce) is requisite to synchronize gradients (detailed in Appendix C). This communication can be costly, especially for models with large hidden dimensions and computing clusters with poor interconnection. Under such conditions, maximizing the overlap between computation and communication is a key to achieving higher throughput. 

We employ eager gradient synchronization \cite{li2021chimera} to overlap the all-reduce overhead with computation. As shown in Figure \ref{EagerSync}(a), an intuitive way for gradient synchronization is to implement the synchronizing step for each stage maintained by the devices after the completion of all the local computations. It is noted that the gradient synchronization for the middle stages (i.e., ${\rm S_6}$\textasciitilde${\rm S_7}$ and ${\rm S_2}$\textasciitilde${\rm S_3}$ in Figure \ref{EagerSync}) is partially overlapped by the computation on the initial and terminal stages (i.e., micro-batch 2 of P1 and micro-batch 4 of P4). To achieve a more profound communication overlap, we initiatively launch allreduce by making use of the bubbles in the pipeline. As shown in Figure \ref{EagerSync}(b), in the case of BitPipe with four pipeline devices, the gradients synchronization of stage5 and stage8 is advanced and overlapped by the bubbles and the following computation.

\begin{figure}[ht]
	\centering
	\includegraphics[width=0.95\columnwidth]{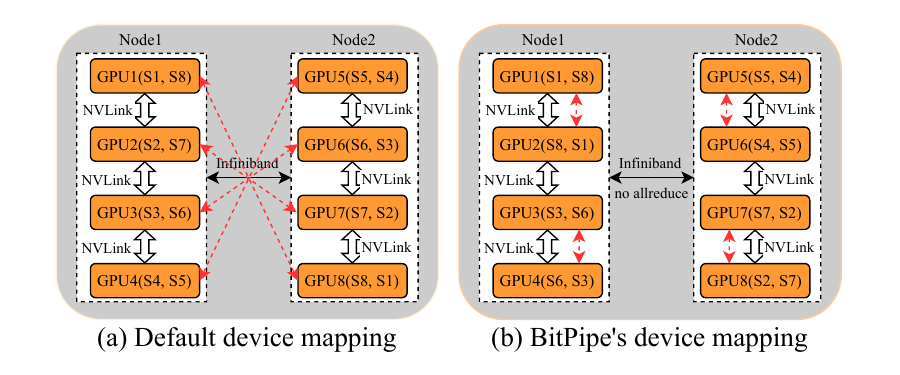}
	\caption{Device mapping for bidirectional pipelines. $\rm S_i$ denotes stage-$\rm i$ for each model replica, and a red dashed double arrow represents an allreduce.}
	\label{devicemapping}
\end{figure}

\begin{figure}[ht]
	\centering
	\includegraphics[width=0.95\columnwidth]{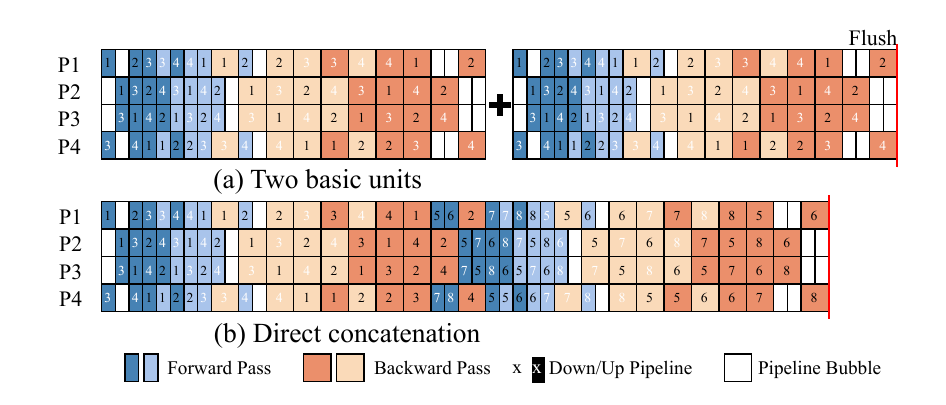}
	\caption{Scale to more than $D$ micro-batches within a training iteration.}
	\label{scale-moreb}
\end{figure}

To improve communication efficiency, we also explore the mapping of pipeline stages onto multiple devices. BitPipe tends to place all replicas of a stage (both in data parallelism and bidirectional pipeline parallelism) into the same server node. This mapping exploits workload characteristics by leveraging the high speed NVLink for heavy gradients synchronization, while using the slow Infiniband for small activations communication, as shown in Figure \ref{devicemapping}.

\subsection{Scale to More Micro-Batches}\label{sec:moreb}

For a large mini-batch, the number of micro-batches in an iteration for each device may be more than $D$ (i.e., $N \textgreater D$), especially when the compute resources are limited. Scaling to a large mini-batch, we first use the schedule of $D$ micro-batches in BitPipe as a basic scheduling unit and scale it by concatenating $K$ ($K$=$N$/$D$) basic units together. Figure \ref{scale-moreb} shows an example with 2$D$ micro-batches per device in a training iteration (i.e., $N$=2$D$), which has two basic units (i.e., $K$=2). The bubbles at the end of the first basic unit can be occupied by the first two forward passes of the second basic unit. The intermediate bubbles can be eliminated by scheduling more forward passes in advance, but at the cost of higher memory usage (detailed discussed in Appendix B).

\section{Experiments}
\subsection{Experimental Setup}

\noindent\textbf{Hardware.} We conduct experiments on a cluster with up to 32 NVIDIA A800 80GB GPUs, where servers are connected by NVIDIA Mellanox 200Gbps HDR Infiniband HCAs and GPUs in a sever are interconnected via NVLink.

\begin{table}[ht]
	\small
	\centering
	\begin{tabular}{ccccc}
		\toprule
		\begin{tabular}[c]{@{}c@{}}
			Models\\(\# Parameters)
		\end{tabular} &
		\begin{tabular}[c]{@{}c@{}}
			\# Layers 
		\end{tabular} &
		\begin{tabular}[c]{@{}c@{}}
			\# Attention \\Heads
		\end{tabular} &
		\begin{tabular}[c]{@{}c@{}}
			Hidden \\Size
		\end{tabular} &
		\begin{tabular}[c]{@{}c@{}}
			Sequence \\Length
		\end{tabular} \\
		\midrule
		BERT-64 (5B) & 64 & 64 & 2560 &512 \\
		GPT-96 (11B) & 96 & 32 & 3072 &1024  \\
		\bottomrule
	\end{tabular}
	\caption{Benchmark models.}
	\label{table-models}
\end{table}

\noindent\textbf{Baselines and Implementation.} We compare BitPipe with four synchronous approaches: (a) DAPPLE of the 1F1B schedule; (b) 1F1B-Int of multiple stages per device;  (c) Chimera of bidirectional pipelines; and (d) MixPipe of bidirectional pipelines with new device mapping. We base our implementation\footnote{\url{https://github.com/wuhouming/BitPipe}} on the open-source Megatron-LM project \cite{narayanan2021efficient}. To be fair, all approaches are implemented in PyTorch with NCCL distributed backend.

\noindent\textbf{Models and Datasets.} We evaluate BitPipe on large transformer-based language models extensively used for natural language processing (NLP) applications, including two variants of BERT and GPT-3, as detailed in Table \ref{table-models}. We use WikiPedia \cite{devlin2018bert} and OpenWebText \cite{peterson2019open} to train the above two models, respectively, and data preprocessing is the same as Megatron-LM \cite{narayanan2021efficient}.

\noindent\textbf{Evaluation Metrics.} We mainly compare the memory footprint and the training throughput, as BitPipe and all baselines are synchronous pipeline approaches. \emph{Throughput} is defined as the number of samples processed per second.

\noindent\textbf{Procedure and Parameter Settings.} We evaluate the pipeline parallelism performance and the parallel scalability combined with data parallelism on 8, 16, and 32 GPUs. The running time of each iteration is recorded after 100 warm-up iterations. All results shown are with mixed precision. We also conduct ablation study and hyperparameter study on BitPipe to investigate the effectiveness of the key components and the impact of hyperparameters.

\subsection{Main Results}

\subsubsection{Pipeline Parallelism Performance.} To evaluate the performance of pipeline parallelism separately, the data parallelism size $W$ and pipeline parallelism size $D$ are set to 1 and 8, respectively. To maximize GPU memory usage, the micro-batch size $B$ is set to 4 for BERT-64 and 1 for GPT-96, respectively. The number of micro-batches $N$ in a mini-batch  scales from $D$ to $2D$ and $4D$, i.e., the mini-batch size $\hat{B}$ scales from 32 to 64 and 128 for BERT-64, or from 8 to 16 and 32 for GPT-96.

\begin{figure}[ht]
	\centering
	\subfigure[8 GPUs]{
		\includegraphics[width=0.47\columnwidth]{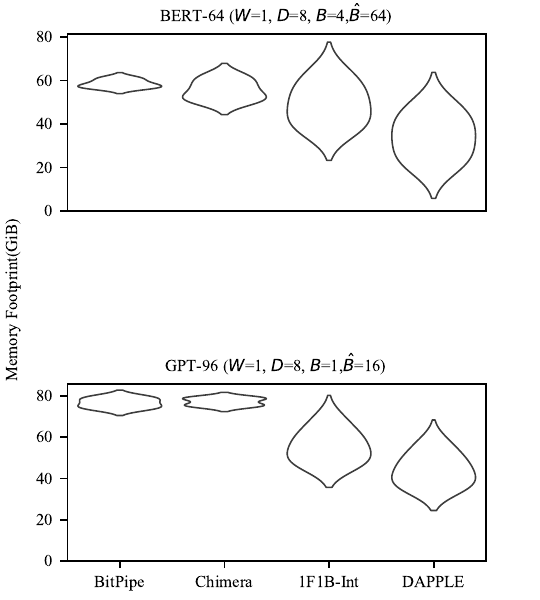}
		\label{MemoryConsumption_1node}
	}
	%\quad
	\subfigure[32 GPUs]{
		\includegraphics[width=0.47\columnwidth]{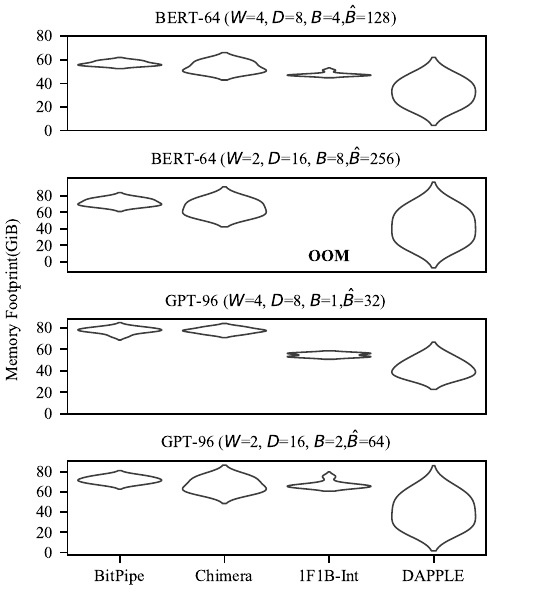}
		\label{MemoryConsumption_4node}
	}
	\caption{Memory footprint distributions.}
	\label{Memory}
\end{figure}

\begin{figure}[ht]
	\centering
	\subfigure[BERT-64]{
		\includegraphics[width=0.45\columnwidth]{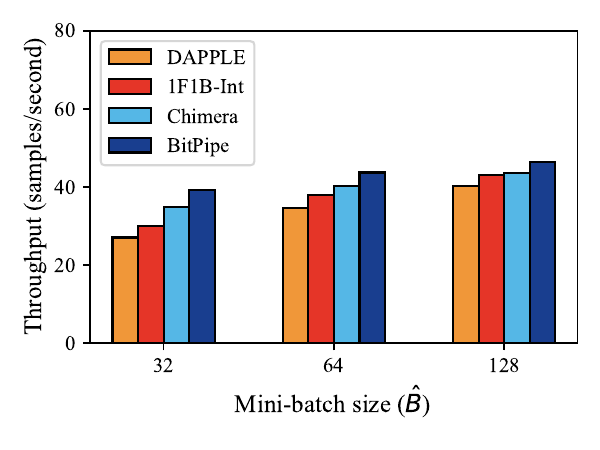}
		\label{TP_1node_bert8}
	}
	\subfigure[GPT-96]{
		\includegraphics[width=0.45\columnwidth]{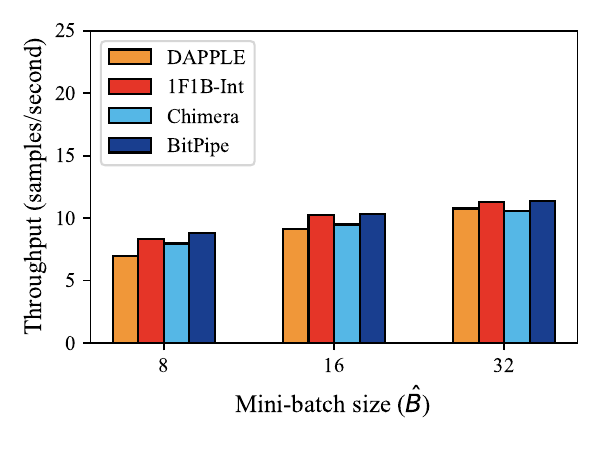}
		\label{TP_1node_gpt}
	}
	\caption{Throughput comparison (only pipeline parallelism) on 8 GPUs.}
	\label{TP_1node}
\end{figure}

\textbf{Memory Footprint.} Figure \ref{MemoryConsumption_1node} presents the memory footprint distribution (including both activations and weights) of training the two models on 8 GPUs. We observe that: (1) 1F1B-Int and DAPPLE display the most imbalanced memory footprint, as they inject different numbers of micro-batches into each device at the beginning of the pipeline, which results in the highest activations memory consumption on the device responsible for the first pipeline stage. (2) Although having higher average memory consumption due to the stashing of two versions of weights and up to $D$ micro-batches’ activations, BitPipe exhibits a narrow and more uniform distribution, which is consistent with the memory analysis in Table \ref{table-methods}. 

\textbf{Throughput.} The throughput comparison of the four pipeline parallelism approaches are displayed in Figure \ref{TP_1node}, and the following tendencies can be discerned: (1) BitPipe consistently outperforms all the baselines across all configurations, as it has the lowest bubble ratio. For BERT-64, BitPipe outperforms DAPPLE, 1F1B-Int, and Chimera by average 1.27$\times$, 1.12$\times$ and 1.09$\times$, respectively. For GPT-96, BitPipe outperforms DAPPLE, 1F1B-Int, and Chimera by average 1.15$\times$, 1.03$\times$ and 1.09$\times$, respectively. (2) The leading edge of BitPipe slows down with the increase in mini-batch size, as BitPipe introduces more P2P communication than the other approaches.

\subsubsection{Parallel Scalability.} To evaluate the parallel scalability of combining with data parallelism, we maintain the same amount of computation per device while incrementally increasing the number of devices used from 8 to 16 and 32.  For each GPU setting, we obtain the best configuration for each approach by grid-searching the space of the parameters ($W$, $D$, and $B$). The number of micro-batches $N$ in a mini-batch equals $D$ by default. Table \ref{table-para-search} presents the search space of parameters and their final choices. 

\begin{table}[ht]
	\small
	\centering
	\setlength{\tabcolsep}{0.8mm}	
	\begin{tabular}{lllcccccccccccc}
		\toprule
		\multirow{3}{*}{Model}&\multirow{3}{*}{Params} &\multirow{3}{*}{\shortstack{Considered \\Values}}& \multicolumn{4}{c}{8 GPUs} & \multicolumn{4}{c}{16 GPUs} & \multicolumn{4}{c}{32 GPUs} \\
		\cmidrule(lr){4-15}
		&&                    & A & I & M & B & A & I & M & B & A & I & M & B\\
		\midrule
		\multirow{3}{*}{\shortstack{BERT\\-64}}&$W$        & \{1, 2, 4, 8\}     & 2&2&1&1 &4&4&4&4&4&8&4&4\\
		&$D$        & \{4, 8, 16\}       & 4&4&8&8 &4&4&4&4&8&4&8&8 \\
		&$B$        & \{1, 2, 4, 8\}     & 4&4&4&4 &4&4&4&4&2&2&4&4 \\
		\midrule
		\multirow{3}{*}{\shortstack{GPT\\-96}}&$W$        & \{1, 2, 4\}     &1&1&1&1 &2&2&2&2&4&4&4&4 \\
		&$D$        & \{8, 16\}       &8&8&8&8 &8&8&8&8&8&8&8&8 \\
		&$B$        & \{1, 2\}     &1&1&1&1 &1&1&1&1&1&1&1&1 \\
		\bottomrule
	\end{tabular}
	\caption{The search space of parameters and their final choices. A stands for DAPPLE, I stands for 1F1B-Int, M stands for MixPipe, and B stands for BitPipe.}
	\label{table-para-search}
\end{table}

\begin{figure}[ht]
	\centering
	\subfigure[BERT-64]{
		\includegraphics[width=0.47\columnwidth]{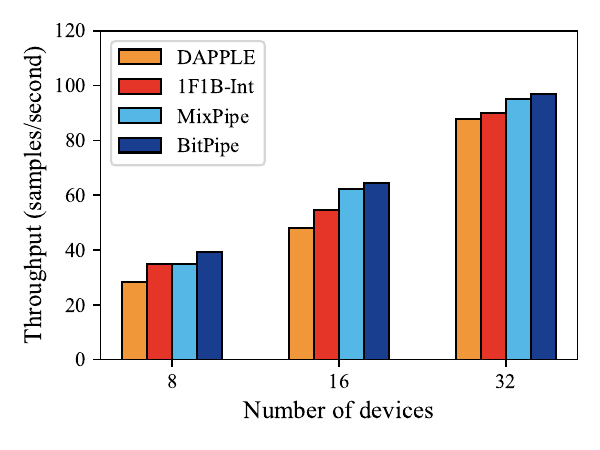}
		\label{bert_weak}
	}
	\subfigure[GPT-96]{
		\includegraphics[width=0.47\columnwidth]{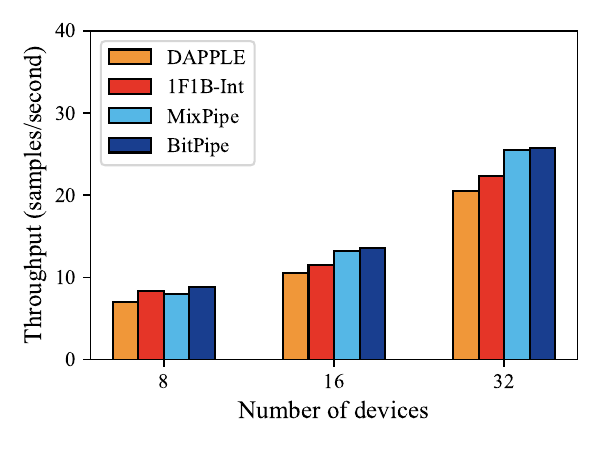}
		\label{gpt_weak}
	}
	\caption{Throughput comparison (combined with data parallelism).}
	\label{Weak_Scale}
\end{figure}
\textbf{Memory Footprint.} Figure \ref{MemoryConsumption_4node} presents the memory footprint distribution in different configurations on 32 GPUs. We observe that: (1) DAPPLE displays the most imbalanced memory footprint that is consistent with that of 8 GPUs. The memory footprint distribution of 1F1B-Int tends to be concentrated, while the peak memory usage increases a lot under larger micro-batch size $B$ and easily leads to OOM (Out of Memory). (2) BitPipe is on par with the SOTA approaches for the peak memory consumption, with a more balanced memory usage among the devices.

\textbf{Throughput.} Figure \ref{Weak_Scale} presents the results and we observe that: (1) BitPipe consistently outperforms all the baselines at all scales. For BERT-64, BitPipe outperforms DAPPLE, 1F1B-Int, and MixPipe by average 1.28$\times$, 1.13$\times$, and 1.06$\times$, respectively. For GPT-96, BitPipe outperforms DAPPLE, 1F1B-Int, and MixPipe by average 1.27$\times$, 1.15$\times$, and 1.05$\times$, respectively. (2) BitPipe have a performance degradation under multi-node settings. This could be due to the synchronization of the two model replicas and the communication overhead caused by the over-fine-grained stages.

\subsection{Ablation Study}

To validate the effectiveness of the V-shaped interleaved schedule and eager gradient synchronization, we compare BitPipe with the following variants:

\noindent\textbf{BitPipe w/o V}: This variant removes the V-shaped interleaved schedule, using the looping schedule of 1F1B-Int.

\noindent\textbf{BitPipe w/o E}: This variant removes the eager gradient synchronization, using default synchronization after the completion of all the local computation.

\begin{table}[ht]
	\small
	\centering
	\begin{tabular}{lcccccc}
		\toprule
		Model & \multicolumn{6}{c}{BERT-64 (5B)} \\
		\# GPU &4 &4 &4 &8 &8 &8   \\
		$D$ &4 &4 &4 &8 &8 &8   \\
		$\hat{B}$ &16 &32 &64&32 &64 &128\\
		\midrule
		w/o V & \underline{19.48} & \underline{22.40} & \underline{24.19} & \underline{38.78} & \underline{43.38} & \underline{46.29}\\
		w/o E & 18.82 & 21.99 & 24.11 & 38.32 & 43.18 & 46.05  \\
		BitPipe & \textbf{19.58} & \textbf{22.54} & \textbf{24.28} & \textbf{39.17}& \textbf{43.69} & \textbf{46.43} \\
		\bottomrule
	\end{tabular}
	\caption{Results of the ablation study. The best results are \textbf{bolded}. The second-best results are \underline{underlined}.}
	\label{table-ablation}
\end{table}

\begin{figure}[ht]
	\centering
	\subfigure[Pipeline parallelism size $D$]{
		\includegraphics[width=0.47\columnwidth]{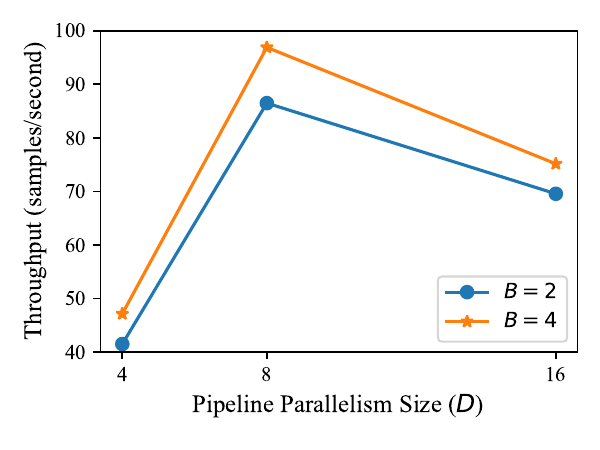}
		\label{Hyper_D_BERT}
	}
	\subfigure[Micro-batch size $B$]{
		\includegraphics[width=0.47\columnwidth]{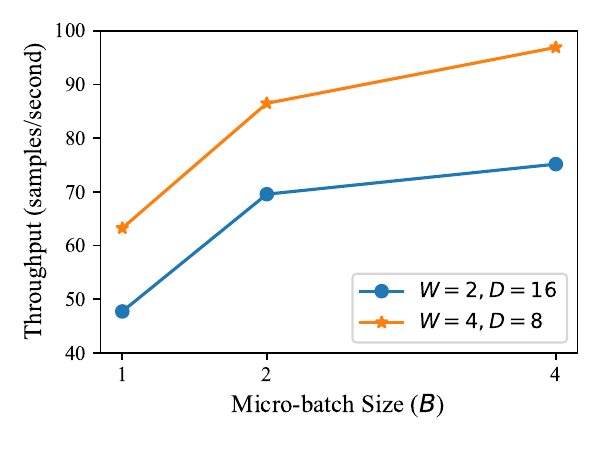}
		\label{Hyper_B_BERT}
	}
	\caption{Results of the hyperparameter study. }
	\label{Hyper_S}
\end{figure}

To negate the influence of cross-node communication, the experiments are conducted on a server node with 8 A800 GPUs fully connected with NVLink. Table \ref{table-ablation} shows the results of variants on BERT-64, and we observe that: (1) BitPipe outperforms the two variants, suggesting  that the V-shaped interleaved schedule and eager gradient synchronization are effective. (2) BitPipe w/o V outperforms BitPipe w/o E, indicating that the eager gradient synchronization plays a greater role than the V-shaped interleaved schedule in reducing/overlapping communication overhead.

\subsection{Hyperparameter Study}

To investigate the impact of pipeline parallelism size $D$ and micro-batch size $B$ on BitPipe, we conduct a hyperparameter study on BERT-64 and 32 GPUs. The mini-batch size $\hat B$ is set to 128. Figure \ref{Hyper_S} presents the results and we observe that: (1) The pipeline parallelism size $D$ is significant to BitPipe, as it controls the communication architecture. Too large or too small $D$ could destroy the mechanism of BitPipe, i.e., using the high-speed NVLink for heavy gradients synchronization and the slow Infiniband for activations communication, resulting in a significant decrease in throughput. (2) BitPipe is also sensitive to the micro-batch size $B$, and the training throughput increases with the increase of $B$. This indicates that when memory and communication are not bottlenecks, larger micro-batch size $B$ should be used to achieve higher throughput.

\section{Conclusions}

In this paper, we propose BitPipe, a bidirectional interleaved pipeline parallelism for accelerating large models training. Specifically, a hybrid scheme of fusing interleaved pipelines with bidirectional pipelines is proposed to reduce the computational time of each single micro-batch and multiply the number of devices executing simultaneously. A V-shaped schedule with eager gradient synchronization is introduced to reduce and overlap the communication between devices. Empirical results of training large language models on up to 32 GPU nodes show that BitPipe significantly improves training throughput and memory balance over the state-of-the-art approaches.

%To reduce the communication cost by exploiting sparsification and quantization, alongside making full use of high-speed network hardware in deep learning is our next step.

\bibliography{aaai25}

\newpage
\appendix
\newpage
\section{Appendix A. Generalize to More Stages}\label{app:A}

\begin{figure}[ht]
	\centering
	\includegraphics[width=\columnwidth]{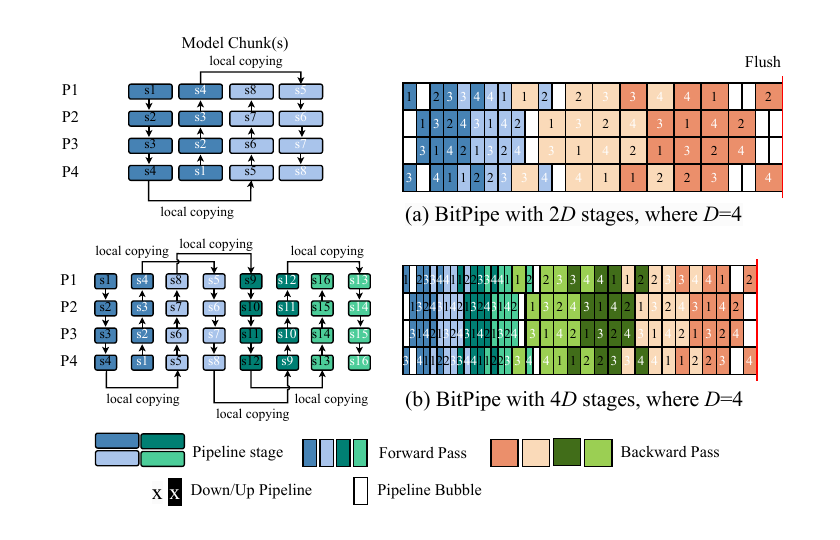}
	\caption{Generalize to more than $2D$ stages per pipeline within a training iteration.}
	\label{scale-mores}
\end{figure}

Although BitPipe can be generalized to incorporate more than two pipelines, which further diminishes the bubbles and balances the activations memory consumption, we do not implement that on account of the expense of weights memory consumption and higher communication overhead. Instead, we choose to generalize the stage number of each pipeline. Theoretically, if each device has $v$ stages (or model chunks), the forward and backward time of a micro-batch for each stage or chunk will be decreased by a factor of $v$. The size of the pipeline bubble is proportional to this time. As depicted in Figure \ref{scale-mores}, the pipeline flushing for the same mini-batch size occurs earlier in BitPipe with more stages per pipeline (Figure \ref{scale-mores}(b)). 

This schedule reduces pipeline bubble size and avoids the extra memory and data-parallel communication overhead associated with generalizing the pipeline number. Nevertheless, it is not without cost: this schedule requires extra P2P communication. Quantitatively, the amount of communication also increases by $v$. Hence, $v$=2 (i.e., a combination of two V-shaped interleaved pipelines with two stages for each pipeline) is the default configuration for BitPipe. We expect that $v$ \textgreater 2 would further improve the performance for future large models featuring a larger pipeline parallelism size.

\begin{figure*}[ht]
	\centering
	\includegraphics[width=0.9\textwidth]{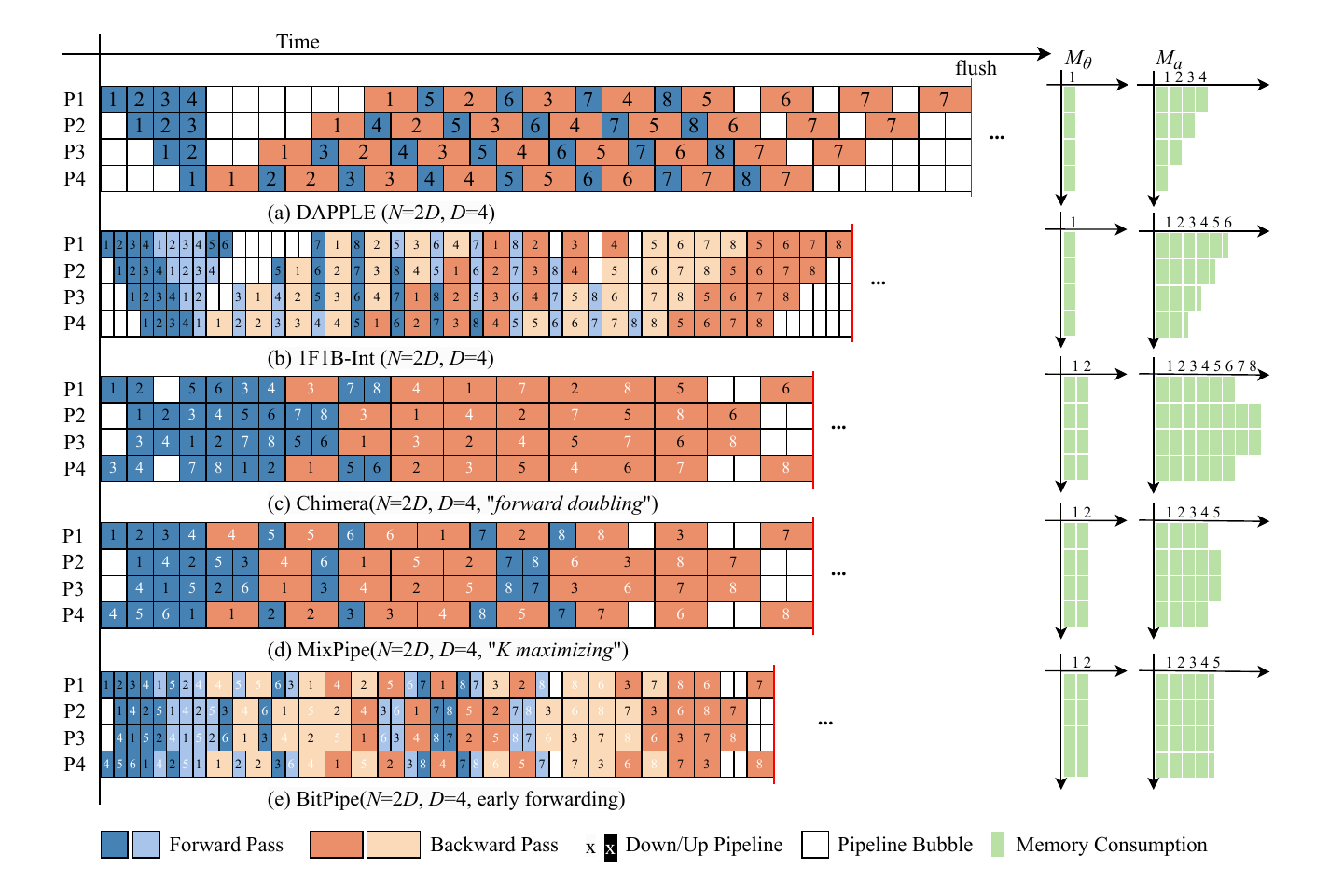}
	\caption{Comparison of the five synchronous approaches, with four pipeline devices ($D$=4) and eight micro-batches ($N$=8) within a training iteration. }
	\label{scale-moreb_2d}
\end{figure*}

\section{Appendix B. Scale to More Micro-Batches with Early Forwarding}\label{app:B}

We introduce an early forward scheduling to balance the workload of forward and backward passes, which removes the intermediate bubbles of direct concatenation, as shown in Figure \ref{scale-moreb_2d}(e). By scheduling the first backward pass in each device as soon as possible, the peak activations memory can be maintained at ((3$D$-3)/2)$M_{\rm a}$, which is lower than that of the scaling methods of Chimera and MixPipe (i.e., 2$D\times$$M_{\rm a}$ for Chimera’s \textit{forward doubling} and ((3$D$-2)/2)$M_{\rm a}$ for MixPipe's $K$ \textit{maximizing}, respectively), and thus has better device utilization. 

\begin{table}[ht]
	\centering
	\resizebox{.95\columnwidth}{!}{
		\begin{tabular}{cc}
			\toprule
			\begin{tabular}[c]{@{}c@{}}
				Pipeline \\Approaches
			\end{tabular} &
			\begin{tabular}[c]{@{}c@{}}
				Communication \\Overhead 
			\end{tabular} \\
			\midrule
			DAPPLE &  $\left(2N+2\left(D-1 \right)\right) \times \frac{message\_size}{W_{\rm inter}}$      \\
			1F1B-Int & $\left(4N + 4\left(D-1 \right)\right) \times \frac{message\_size}{W_{\rm inter}}$    \\
			Chimera & $\left(2N + 2\left(D-1 \right)\right) \times \frac{message\_size}{W_{\rm inter}} + \frac{M_{\rm grad}}{W_{\rm intra}}$            \\
			\textbf{BitPipe} & $\left(4N + 4\left(D-1 \right)\right) \times \frac{message\_size}{W_{\rm inter}}+ \frac{M_{\rm grad}}{W_{\rm intra}}$   \\
			\bottomrule
	\end{tabular}}
	\caption{Communication overhead of different approaches, with $D$ pipeline devices and $N$ micro-batches within a training iteration. The message size of per NCCL call ($message\_size$ ) is calculated as $2 {\rm Bytes}\times$$B\times$$S\times$$H$, where $B$ is the micro-batch size, $S$ the sequence length, and $H$ the hidden size. $W_{\rm inter}$ and $W_{\rm intra}$ are the communication bandwidth between and within the compute server, respectively.}
	\label{T_communication}
\end{table}

\begin{table}[ht]
	\small
	\centering
	\begin{tabular}{lccccccc}
		\toprule
		Model & \multicolumn{3}{|c}{BERT-64 (5B)} & \multicolumn{2}{|c}{GPT-96 (11B)}  \\
		\# GPU &\multicolumn{1}{|c}{32} &32 &32 &\multicolumn{1}{|c}{32} &32  \\
		$D$ &\multicolumn{1}{|c}{4} &8 &16 &\multicolumn{1}{|c}{8} &16  \\
		$\hat{B}$ &\multicolumn{1}{|c}{128} &128&128 &\multicolumn{1}{|c}{32}&32 \\
		\midrule
		DAPPLE & \multicolumn{1}{|c}{85.10}& \textbf{87.79} & 82.26 & \multicolumn{1}{|c}{\textbf{20.51}} & 19.44 \\
		1F1B-Int & \multicolumn{1}{|c}{\textbf{89.88}} & 85.90 & 60.32 &  \multicolumn{1}{|c}{\textbf{22.30}} & 19.51   \\
		MixPipe & \multicolumn{1}{|c}{38.59} & \textbf{95.10} & 76.97 & \multicolumn{1}{|c}{\textbf{25.45}} & 24.06  \\
		BitPipe & \multicolumn{1}{|c}{47.15} & \textbf{96.90} & 75.16 & \multicolumn{1}{|c}{\textbf{25.70}} & 22.22\\
		\bottomrule
	\end{tabular}
	\caption{Performance tuning for different pipeline parallelism approaches on 32 GPUs. The best result for each approach is \textbf{bolded}.}
	\label{table-tuning}
\end{table}

It can be observed that BitPipe incurs ($D$-2)/2 bubbles (i.e., ($D$-2)/4 bubbles in the forward passes and ($D$-2)/4 bubbles in the backward passes). The total amount of time spent in the pipeline bubble $t_{\rm {pb}}$ and the ideal processing time for the mini-batch $t_{\rm {id}}$ can be calculated as follows:
\begin{equation}
	\begin{aligned}
		t_{\rm {pb}} &= \frac{D-2}{4}\cdot(t_{\rm f}+t_{\rm b}) \\
		t_{\rm {id}} &= N\cdot (t_{\rm f}+t_{\rm b})
	\end{aligned}
\end{equation}
where $t_{\rm f}$ and $t_{\rm b}$ are the time to execute a single micro-batch’s forward and backward pass, respectively. Given the assumption that $t_{\rm b}$ is two times of $t_{\rm f}$, the bubble ratio of BitPipe is:
\begin{equation}
	{bubble\_ratio}=\frac{t_{\rm {pb}}}{t_{\rm {id}}+t_{\rm {pb}}}=\frac{D-2}{4N+D-2}
\end{equation}
which is lower than that of BitPipe with direct concatenation (i.e., ($D$-2)/(3$N$+$D$-2)).

Therefore, BitPipe with early forwarding not only has the least bubbles but also exhibits a more balanced and lower peak memory footprint.

\section{Appendix C. Communication Overhead}\label{app:C}

The communication overhead of pipeline parallelism in one iteration can be obtained by multiplying the time of a single communication by the number of communications. Table \ref{T_communication} presents the communication overhead of the four pipeline parallelism approaches. For DAPPLE and 1F1B-Int, the communication overhead is primarily the P2P communication to transfer the intermediate activations and gradients between pipeline stages. 1F1B-Int doubles the number of pipeline stages, and the communication overhead also doubles accordingly. As Chimera and BitPipe combine bidirectional pipelines together, collective communication (i.e., allreduce) is requisite to synchronize the weight gradients. BitPipe has the largest communication overhead as it doubles the number of pipeline stages.

In addition, the communication bandwidth between and within the server node will also affect the communication overhead, especially when there is a bottleneck in the communication bandwidth between the clusters. This is why BitPipe's leading advantage slows down as the number of devices increases. Table \ref{table-tuning} presents the performance tuning of the four pipeline parallelism approaches for BERT-64 and GPT-96 with mini-batch size 128 and 32, respectively. The training throughput of pipeline parallelism size 8 is relatively better than that of other parallelism size, as it achieves the best compromise between bubbles and communication overhead.

To extend BitPipe, exploiting sparsification and quantization to reduce the communication cost, alongside making full use of high-speed network hardware are possible directions.
%\fi
\end{document}